\documentclass{article}

\usepackage{iclr2027_conference,times}
\iclrfinalcopy
\usepackage[utf8]{inputenc}
\usepackage[T1]{fontenc}
\usepackage{hyperref}
\hypersetup{hidelinks}
\usepackage{url}
\usepackage{booktabs}
\usepackage{amsfonts}
\usepackage{amsmath}
\usepackage{amssymb}
\usepackage{amsthm}
\usepackage{mathtools}
\usepackage{nicefrac}
\usepackage[expansion=false]{microtype} 
\usepackage{xcolor}
\usepackage{multirow}
\usepackage{array}
\usepackage{graphicx}
\usepackage{subcaption}
\usepackage{tcolorbox}
\usepackage{cleveref}
\usepackage{enumitem}
\usepackage{siunitx}
\usepackage{longtable}
\usepackage{float}
\usepackage{wrapfig}
\usepackage{tikz}
\usepackage{pgfplots}
\usetikzlibrary{shapes.geometric, arrows.meta}
\definecolor{poolgreen}{RGB}{185, 235, 195}
\definecolor{qiborder}{RGB}{100, 180, 100}
\definecolor{expertblue}{RGB}{145, 215, 215}

\newcommand{\R}{\mathbb{R}}
\newcommand{\E}{\mathbb{E}}
\newcommand{\St}{\mathrm{St}}
\newcommand{\SPD}{\mathrm{SPD}}

\newcommand{\upper}{\mathrm{upper}}
\newcommand{\softmax}{\mathrm{softmax}}

\newcommand{\bacc}{\mathrm{bAcc}}
\newcommand{\gap}{\mathrm{Gap}}

\DeclareMathOperator*{\argmax}{arg\,max}

\theoremstyle{plain}
\newtheorem{proposition}{Proposition}
\newtheorem{lemma}{Lemma}
\newtheorem{corollary}{Corollary}
\theoremstyle{definition}
\newtheorem{definition}{Definition}
\theoremstyle{remark}

\title{%
Escaping the Capacity Ceiling: \newline
Routing on the Stiefel Manifold \newline
for Bilinear SPD Layers
}

\author{%
  Isabella Costa Maia\textsuperscript{1}, \,
  Salem Said\textsuperscript{2}, \,
  Pedro L.~C. Rodrigues\textsuperscript{2, 3}, \,
  Marco Congedo\textsuperscript{1} \\[6pt]
  \textsuperscript{1}GIPSA-lab, Univ.\ Grenoble Alpes, CNRS, Grenoble-INP \\
  \textsuperscript{2}Univ.\ Grenoble Alpes, CNRS, Grenoble INP, LJK \\
  \textsuperscript{3}Univ.\ Grenoble Alpes, Inria \\[4pt]
  \texttt{isabella.costa-maia@grenoble-inp.fr} \\
  \texttt{salem.said@univ-grenoble-alpes.fr} \\
  \texttt{pedro.rodrigues@inria.fr} \\
  \texttt{marco.congedo@gipsa-lab.grenoble-inp.fr}
}

\begin{document}
\maketitle

\begin{abstract}
Deep networks on the symmetric positive-definite (SPD) manifold promise expressive representations by encoding data geometry as an inductive bias, but stacking BiMap layers with the standard ReEig nonlinearity often adds no capacity: on real, preconditioned EEG data, ReEig rarely activates, so the stack behaves as a single layer at any depth. In the worst case, when domains share no discriminative directions, we prove a single filter has a capacity ceiling, so it cannot fully align every domain at once. To overcome that, we propose SCAP (Stiefel Cross-Attention Pool), a layer implementing a family of Stiefel filters by combining a pool of $K$ experts into a sample-specific bilinear map via cross-attention. We show that it matches a per-domain filter bank to first order with fewer experts than domains when domain-optimal filters span few directions near a shared tangent-space basepoint; in the worst case, its alignment empirically stays nearly flat as domains grow, escaping the fixed-filter ceiling. Naively trained, however, this routing can collapse to a fixed filter; we diagnose why and adapt three mechanisms to mitigate it. SCAP significantly improves balanced accuracy over fixed-filter SPDNet on all five cross-domain EEG motor-imagery datasets, and matches or exceeds three domain-adaptive baselines on four out of five.
\end{abstract}

\section{Introduction}
\label{sec:intro}

Brain-computer interfaces (BCI) built on electroencephalography (EEG) decode a subject's intent from brain electrical signals to perform tasks such as inferring motor intention and spelling. Since their inception in \cite{barachant2010riemannian}, classifiers based on Riemannian geometry have become the state-of-the-art (SotA) in decoding EEG signals \citep{chevallier2024moabb}: instead of considering the data in a flat Euclidean space, spatial covariance matrices started being used as data descriptors, carrying the geometry of the curved space to which they belong and reaching unprecedented performance in the field. More specifically, covariance matrices belong to the Symmetric Positive Definite (SPD) manifold, a curved and differentiable space where gradient-based methods can be used. Since the introduction of SPDNet by \cite{huang2017riemannian}, deep learning methods have been applied to this realm by learning manifold-constrained parameters that preserve the geometry of the data throughout the network, building that geometry into the model as an inductive bias. One of its canonical layers, BiMap, accounts for the only manifold-valued parameter learnt in the network. It performs a single dimensionality-reduction bilinear operation on all data points, that is, a single spatial filter to fit all data (\Cref{sec:preliminaries}).

Currently, most deep SPD models use SPDNet as backbone, extended with additional BiMap-ReEig blocks \citep{wang2022dreamnet,wang2023uspdnet}, attention mechanisms \citep{wang2023selfattention}, domain-normalization layers \citep{brooks2019riemannian,kobler2022spd}, or alternative front-ends coupling SPDNet with learned space-time, spectral, or phase-space features \citep{kobler2022spd,paillard2025green, carrara2025geometric} (\Cref{sec:related-work}). 

Despite the promise behind SPDNet-based deep networks, the literature consistently reports a performance gap: these models do not outperform shallow Riemannian methods (Riemannian classifiers not based on gradient descent, such as Minimum Distance to Mean (MDM) \citep{barachant2010riemannian,barachant2012multiclass}) and, unlike stacking convolutional layers, do not scale with depth \citep{wang2022dreamnet,wang2023uspdnet, wang2023selfattention}, so networks in practice stay shallow (\Cref{sec:related-work}). As we show in \Cref{sec:depth-closure}, this is not only an empirical pattern: concurrent theoretical work \citep{oswald2026expressivity} proves that, architecturally, a rank-reducing BiMap-ReEig stack of any depth collapses to a structure with at most one active nonlinearity, since every layer past the first is guaranteed to fall inside the activation's own identity region. Combined with the redundancy of ReEig activation previously observed in practice \citep{kobler2022spd,wilson2025deepriemannian}, which we confirm on real EEG data (\Cref{app:reeig-activation}), the network's manifold-valued layers collapse, in practice, to a single fixed linear transform. EEG data is extremely complex, suffering not only from low signal-to-noise ratio but also from high variability: whether we consider inter-subject or inter-session data (data acquired from the same subject but at different moments and electrode locations), covariate shift is intrinsic to the data. Therefore, finding solutions for making these  networks more expressive is of paramount importance.  

\begin{figure}[H]
  \centering
  \resizebox{0.72\linewidth}{!}{
%
%
%
%

\begin{tikzpicture}[
    every node/.style={font=\fontsize{9}{11}\selectfont\sffamily},
]

    
    \draw[rounded corners=20pt, thick, draw=black!80]
        (-3.5, -3.0) rectangle (3.5, 3.0);
    \node[anchor=north west, font=\fontsize{14}{16}] at (-3.4, 3.6)
        {\bfseries SCAP layer};

    \fill[poolgreen, rounded corners=14pt]
        (-0.8, -1.6) rectangle (3.2, 2.5);
    \node[anchor=south] at (1.2, 2.45)
        {\textbf{Stiefel Pool}};

    \node (qi) [draw=qiborder, fill=white, very thick,
                inner sep=8pt, rounded corners=4pt,
                font=\fontsize{16}{18}\selectfont\sffamily]
        at (-2.5, 0.6) {$q_i$};

    \node (ek) [regular polygon, regular polygon sides=3,
                fill=white, minimum size=0.9cm, inner sep=0pt]
        at (0.5, 1.6) {};
    \node at (0.5, 1.52) {$\mathcal{E}_j$};

    \node (wk) [rectangle, fill=expertblue,
                minimum width=0.55cm, minimum height=1.0cm]
        at (1.3, 0.9) {};
    \node at (1.3, 0.9) {$W_j$};

    \node (ek2) [regular polygon, regular polygon sides=3,
                 fill=white, minimum size=0.5cm, inner sep=0pt]
        at (1.9, -0.2) {};
    \node (wk2) [rectangle, fill=expertblue,
                 minimum width=0.28cm, minimum height=0.5cm]
        at (2.3, -0.45) {};

    \node (ek3) [regular polygon, regular polygon sides=3,
                 fill=white, minimum size=0.5cm, inner sep=0pt]
        at (0.1, -0.85) {};
    \node (wk3) [rectangle, fill=expertblue,
                 minimum width=0.28cm, minimum height=0.5cm]
        at (0.45, -1.1) {};

    \node[white] at (1.1, -0.65) {$\cdots$};

    \node at (0, -2.3)
        {$W_i = \mathcal{R}_{W_{\text{base}}}
         \!\left( \sum_j \alpha_{i,j}
         \Pi_{W_{\text{base}}}(W_j) \right)$};

    \draw[-{Stealth[scale=1.3]}, thick]
        (-5.6, -0.4) -- (-2.5, -0.4) -- (qi.south);
    \node[anchor=south] at (-4.6, -0.4)
        {$X_i \in \mathrm{SPD}(n)$};
    \node[anchor=north] at (-4.6, -0.4)
        {$d_i \in \{1,\dots,m\}$};

    \draw[-{Stealth[scale=1.3]}, thick]
        (3.5, -0.4) -- (6.0, -0.4);
    \node[anchor=south] at (4.65, -0.4)
        {$Y_i = W_i^T X_i W_i$};

    \draw[-{Stealth[scale=1.1]}]
        (qi.east) to [bend left=20] (0.9, -0.4);
    \node[inner sep=2pt] at (-2.0, -0.75)
        {$\begin{array}{c}
           \alpha_i = \\[2pt]
           \mathrm{softmax}\!\left(\dfrac{q_i\mathcal{E}^\top}{\sqrt{m}}\right)
         \end{array}$};

    \draw[densely dotted, -{Stealth}]
        (qi.north) to [bend left=25] (0.7, 3.45)
        node[right] {Query};

    \draw[densely dotted, -{Stealth}]
        (ek.north east) to [bend left=15] (4.1, 2.4)
        node[right] {Key};

    \draw[densely dotted, -{Stealth}]
        (wk.east) to [bend left=15] (4.15, 1.1)
        node[right, align=left] {Stiefel\\Expert};

\end{tikzpicture}}
  \caption{SCAP layer. Cross-attention routes each input to a
           sample-specific Riemannian barycenter of $K$ Stiefel experts.}
  \label{fig:scap-layer}

\end{figure}

Informally, a single BiMap filter that preserves the structure of the data cannot specialize toward any one domain (distinct subject, session, or acquisition setup) in particular, as the $m$ domains compete for the same fixed capacity. This limitation is formalized in \Cref{sec:expressivity-gap}.

We propose a layer to raise the filter capacity ceiling (\Cref{def:overlap,lem:pigeonhole}) by letting it depend on the input. We formalize this as a family of Stiefel filters routed by cross-attention (\Cref{def:routed-family}; \Cref{fig:scap-layer} shows the resulting layer) and show it can match a dedicated per-domain filter bank, to first order, using as few as $r{+}1\le m$ experts, where $r$ is the number of independent directions along which the $m$ domains' optimal filters vary. $r$ can fall well below $m-1$ when that variation is driven by a small number of factors shared across subjects and sessions, rather than by each domain varying independently of the others (\Cref{app:expert-compression}). This match holds, with small approximation error, when these optimal filters all lie close to some common filter, which SCAP learns as the anchor around which it blends its experts. Stiefel Cross-Attention Pool (SCAP) realizes this routed family as a mixture of experts interpolated on the Stiefel manifold (\Cref{sec:scap}), with three structural mechanisms mitigating routing collapse in practice. We address the cross-domain setting in which, as in Riemannian domain-adaptation methods \citep{kobler2022spd,rodrigues2018riemannian}, domain identity is known at test time: a single domain-aware model is trained on a cohort of subject-session recordings (domains), each appearing, under different trials, in the training, validation, and test splits.

In summary: \textbf{1. Diagnosis.} A semi-orthogonal BiMap$\to$ReEig stack collapses to one BiMap$\to$ReEig$\to$BiMap block at any depth \citep{oswald2026expressivity}; we show ReEig never activates there on many real EEG data either, so the stack collapses in practice to a single fixed filter (\Cref{sec:depth-closure}). \textbf{2. Theory.} A provable capacity ceiling on any fixed filter serving multiple non-overlapping domains, and a routed family that matches a per-domain filter bank to first order with fewer experts than domains count when domain-optimal filters vary along $r<m-1$ directions near a shared anchor (\Cref{prop:expert-compression}). \textbf{3. Layer.} SCAP, a drop-in BiMap replacement realizing that family as a Stiefel-manifold expert mixture, with three structural mechanisms against routing collapse (\Cref{sec:scap}). \textbf{4. Evidence.} Validation on theory-matched synthetic data and five real EEG datasets against SPDNet and established cross-domain baselines (\Cref{sec:experiments}).

\section{Related Work}
\label{sec:related-work}
\paragraph{SPDNet.}
\citet{huang2017riemannian} introduced SPDNet, the first deep network with manifold-preserving operations for SPD data, built from three non-Euclidean layers: BiMap (discussed above), ReEig, a non-linear activation similar to ReLU \citep{agarap2018deep} but applied to covariance eigenvalues, and LogEig, a tangent-space mapping that linearizes the data at the identity before a fully-connected (FC) classification layer:
\begin{equation}
    \label{eq:SPDNet}
    \texttt{BiMap $\rightarrow$ ReEig $\rightarrow$ LogEig $\rightarrow$ FC}
\end{equation}

\paragraph{Representational expressivity in Riemannian networks.} 

The performance plateau or degradation observed in stacking BiMap $\to$ ReEig layers in deep SPD models is not a new observation in isolation, and each work below attributes it to structural information loss compounding repeated rank-reducing compression. \citet{wang2022dreamnet}'s DreamNet enriches the deep architecture with a Riemannian autoencoder and residual shortcuts on top of an SPDNet backbone; \citet{wang2023uspdnet}'s U-SPDNet adds progressive compression-decompression via encoder-decoder skip connections; both still observe degradation beyond a certain stacking depth, likely tied to rank deficiency in decompression. \citet{wang2023selfattention} instead applies self-attention across the sequence of SPD representations from different BiMap$\to$ReEig levels of a stacked SPDNet backbone. In our solution, rather than summarizing multiple stacked representations, SCAP uses cross-attention to select a single, input-specific combination of filters applied as one bilinear step.

The argument we present in \Cref{sec:depth-closure} is complementary to these works. Rather than targeting increased depth, we address the representational expressivity gap in SPDNets by giving more representational capacity to its primary learnable layer through input-conditioned selection from a filter pool.

\paragraph{SPD domain adaptation: RPA, BN, DSBN.} 
SCAP is primarily conceived as an expressivity fix for SPDNet-like models, proposing a replacement for BiMap; since domain information is used in it, it can also be read as a domain-adaptive solution. We therefore evaluate it against standard SPDNet as well as three domain-adaptive baselines: Riemannian Procrustes Analysis (RPA) \citep{rodrigues2018riemannian}, Batch-Normalization (BN) \citep{brooks2019riemannian}, and Domain-Specific Batch Normalization (DSBN) \citep{kobler2022spd}.

RPA is a fixed, non-differentiable preprocessing step: a per-domain recentering and bilinear map is computed once, prior to classification trianing, unlike SCAP's per-input, end-to-end-learned filters. BN and DSBN are deep-network layers that learn recentering and stretching parameters downstream of ReEig; since SCAP replaces the upstream BiMap, they act on different manifold transforms and are in principle composable with SCAP rather than direct competitors. We compare against all three regardless, to isolate SCAP's expressivity gain against standard domain-adaptive baselines. A more recent BN variant, ARMAGNAC \citep{gallet2026armagnac}, has since appeared; we leave a direct comparison to future work.

\paragraph{Prompt pools and mixture-of-experts routing.}
Cross-attention has previously been used in the literature to collect input-specific information from a shared pool. In L2P \citep{wang2022learning}, a query built from Vision Transformer (ViT) \citep{dosovitskiy2020vit} features selects the top-$p$ prompts from a learned pool, which are concatenated to the input to enrich the context of another ViT. Such a concatenation is not available to us: for SPD data, appending queried values directly to the input would break the manifold structure. This is why, unlike L2P, SCAP formulates the input-enriching operation as a bilinear map, a transformation under which the SPD manifold is closed. Its filter is obtained by interpolating experts' tangent projections with the router's weights, rather than by selecting a discrete subset of experts as L2P does. To keep the expert pool from collapsing to a shared mean, SCAP updates keys sparsely, and only the query's best-matching key receives a training signal per sample \citep{shazeer2017outrageously,fedus2022switch}. Closer to our setting, MUSUME \citep{yamamoto2021subspace} also learns several Stiefel filters for EEG SPD data, but as an input-independent method not using gradient descent and built on top of MDM classification in each fixed subspace.
\section{Preliminaries}
\label{sec:preliminaries}

This section defines the main data spaces considered in the paper, revisits briefly some definitions on Riemannian manifolds and define formally the BiMap layer.

\paragraph{Riemannian manifolds.} A Riemannian manifold $\mathcal{M}$ is a smooth topological space that generalizes the notions of distances and angles to curved geometries, and that can be locally linearized by projection onto a tangent space $T_X\mathcal{M}$, with $X \in \mathcal{M}$.

In the following, we consider $X \in \SPD(n)$ the set of preconditioned EEG covariance matrices occupying $\SPD(n) = \{X \in \R^{n\times n} : X = X^\top,\ X \succ
0\}$, the manifold of $n\times n$ symmetric positive-definite matrices.

We also consider the Stiefel manifold $\St(n,k) = \{W \in
\R^{n\times k} : W^\top W = I_k\}$ of $k$-frames in $\R^n$, the space of matrices with $k$ orthonormal columns, each specifying a $k$-dimensional subspace of $\R^n$ together with an orthonormal basis for it.
We additionally require moving between nearby points of $\St(n,k)$: given $W, W' \in \St(n,k)$, the tangent projection $\Pi_W(W') = W' - W\,\mathrm{sym}(W^\top W')$ maps $W'$ into the tangent space at $W$, and a retraction $\mathcal{R}_W$ maps back onto $\St(n,k)$; \Cref{app:tangent-retraction} gives the full construction used throughout the paper. 

\paragraph{BiMap Layer.} 
\citet{huang2017riemannian} implements the following bilinear map: for $W \in \St(n,k)$, $f_W(X) = W^\top X W$ maps $\SPD(n) \to \SPD(k)$. Orthonormality of $W$'s columns makes $f_W$ a subspace selection, re-expressing $X$ in the $k$-dimensional subspace $W$ spans. 

The table containing the full notation carried over the paper is present in Appendix \ref{app:notation}.

\section{The Expressivity Gap}
\label{sec:expressivity-gap}

\subsection{A fixed filter's overlap ceiling}
\label{sec:pigeonhole}

Consider a single BiMap filter $W \in \St(n,k)$ shared across $m$ recording
domains (different subjects, sessions, or acquisition setups), where domain $d$'s discriminative signal lives in its own $k$-dimensional subspace $U_d \in \St(n,k)$. We study the case where these subspaces are mutually orthogonal, $U_d^\top U_{d'} = 0$ for $d \ne d'$: domains that are, in a precise sense, maximally distinct from one another. This is a deliberate idealization, the least favorable case for a single filter, since any overlap between domains could only make a single filter's job easier.

How well can one fixed $W$ align with all $m$ of these subspaces at once?

\begin{definition}[Per-domain subspace overlap]
\label{def:overlap}
Define $s_d(W) := \|U_d^\top W\|_F^2 \in
[0,k]$ — a continuous count of how many of the filter's $k$ dimensions
fall inside domain $d$'s subspace. Each column $w_j$ of $W$ contributes
$\|U_d^\top w_j\|^2 \in [0,1]$, the squared cosine of the angle between
$w_j$ and domain $d$'s subspace: $1$ if $w_j \in U_d$ exactly, $0$ if
$w_j \perp U_d$. Summed over $W$'s $k$ columns, $s_d = k$ means every
retained dimension lands inside $U_d$; $s_d = 0$ means none do.
\end{definition}

\begin{lemma}[Overlap ceiling]
\label{lem:pigeonhole}
For any $W \in \St(n,k)$ and $m$ mutually orthogonal domain subspaces
$U_1,\ldots,U_m \in \St(n,k)$,
\[
  \sum_{d=1}^m s_d(W) \;\le\; k
  \quad\Longrightarrow\quad
  \frac{1}{m}\sum_{d=1}^m s_d(W) \;\le\; \frac{k}{m}.
\]
\end{lemma}
\begin{proof}
Fix a column $w_j$ of $W$; it is a unit vector. Mutual orthogonality lets
us stack $U_1,\ldots,U_m$ into a single matrix with orthonormal columns,
so the same argument that bounds any one $\|U_d^\top w_j\|^2$ by $\|w_j\|^2
= 1$ applies to their sum: $\sum_d \|U_d^\top w_j\|^2 \le 1$. Summing this
over $W$'s $k$ columns gives $\sum_d s_d(W) = \sum_j \sum_d \|U_d^\top
w_j\|^2 \le k$.
\end{proof}

Read simply, a filter as BiMap has $k$ units of subspace overlap
to spend, and spending more on one domain leaves strictly less
for the rest. Averaged over $m$ domains, that ceiling is $k/m$, meaning that doubling the domain count halves the average filter coverage, independent of training procedure and data volume. In practice, some of the $k$ dimensions can be spent on directions that help no domain at all, so resulting capacity of a learned filter might also undershoot.

\begin{figure}[h]
\centering
\includegraphics[width=\linewidth]{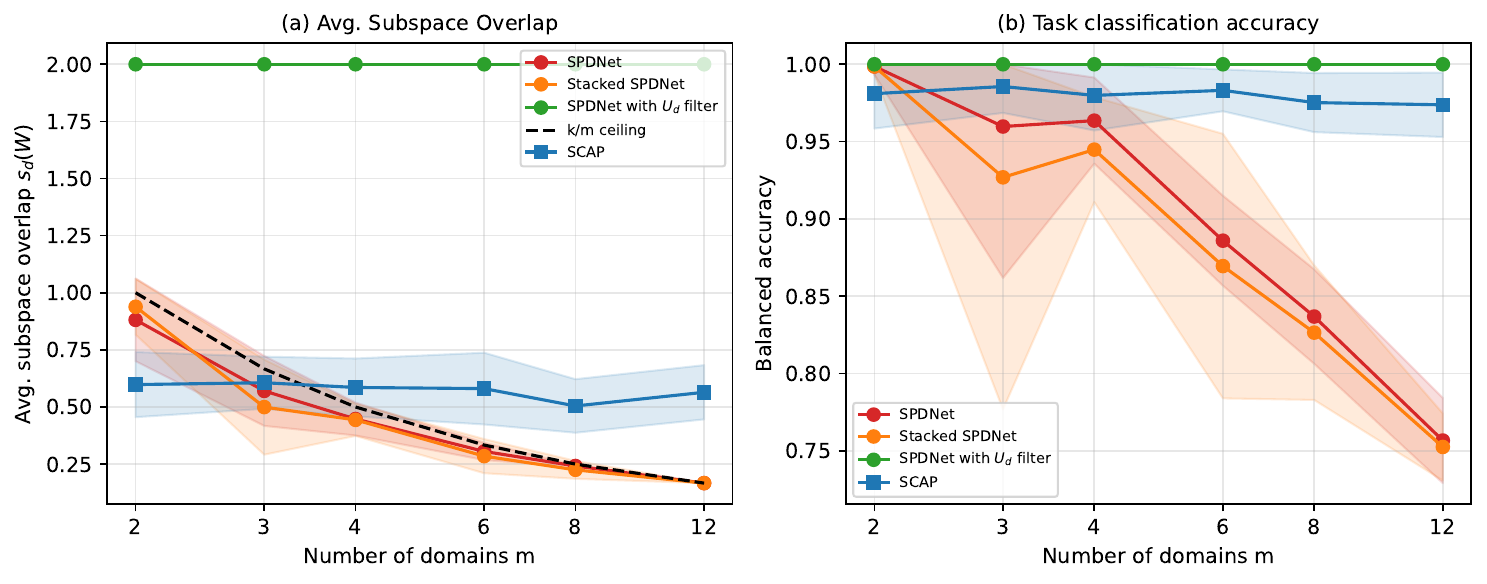}
\caption{Synthetic validation ($n{=}24$, $k{=}2$) with shaded standard deviations. \textbf{(a)} Average subspace overlap vs.\ number of domains $m$, for SPDNet, Stacked SPDNet (4 stacked layers), SPDNet using $W_d=U_d$ (oracle), and SCAP with $K=m$, against the $k/m$ ceiling. \textbf{(b)} Balanced accuracy for the same models.}
\label{fig:overlap-ceiling}
\end{figure}

\Cref{fig:overlap-ceiling} reproduces the toy experiment of \Cref{sec:pigeonhole} and full construction details can be found in \Cref{app:toy-experiment}. The shared filters track the $k/m$ ceiling closely, as \Cref{lem:pigeonhole} predicts; the oracle (SPDNet with ground-truth filter $U_d$) sits flat at $k$, since a dedicated filter per domain never has to compromise. SCAP, with $K{=}m$ experts, keeps its per-domain overlap roughly constant across domain counts, where every fixed filter decays as $k/m$, and exceeds the ceiling by a growing margin from $m{=}4$ on. Importantly, this setting is also far from the regime our theory favors for SCAP: no single tangent-space anchor, the basepoint around which SCAP blends its experts, can lie close to most mutually orthogonal domain subspaces (formalized as $W_\text{base}$ in \Cref{sec:scap}), which is where the first-order guarantee of \Cref{prop:expert-compression} is least informative (\Cref{app:expert-compression}). SCAP's gap to the oracle's overlap plausibly reflects its objective: it is trained for classification, not overlap, and stays within 3 points of the oracle's accuracy as $m$ grows from 2 to 12 (98\%$\to$97\%), while fixed filters degrade from 100\% to 75-76\%.

Note also that Stacked SPDNet has no active
nonlinearity between layers by construction, matching
\Cref{sec:depth-closure}'s linear-stack setting exactly; as we'll show in section \ref{sec:depth-closure}, and as demonstrated in appendix \ref{app:reeig-activation} on real EEG data, that idealization is often close to what a real ReEig-bearing stack does in practice too, since ReEig frequently never activates under standard preconditioning.

\subsection{Depth alone cannot raise the ceiling}
\label{sec:depth-closure}

It is easy to conclude that composing multiple BiMap layers would not increase the network's expressivity either: the composition of Stiefel filters is itself a Stiefel filter ($W_1^\top W_1 = W_2^\top W_2 = I \Rightarrow (W_1W_2)^\top(W_1W_2)=I$, and by induction at any depth $L$),
so a depth-$L$ stack remains one member of the fixed-filter class $\mathcal{F}_k=\{f_W: W\in\St(n,k)\}$, not a larger one, with the same $k/m$ ceiling as \Cref{sec:pigeonhole}.

ReEig is the operation in the standard SPDNet pipeline that can break this closure if activated. Stacking BiMap$\to$ReEig blocks seems then natural for expressivity gain. However, \citet{oswald2026expressivity} demonstrates that for any nonlinear activation whose image lies inside its own identity region (as is the case for ReEig, a shifted ReLU on eigenvalues, with image $[\epsilon,\infty)$) composing it with a rank-reducing congruence stack is architecturally degenerate. By the Poincar\'e separation theorem, a semi-orthogonal congruence transformation only shrink a matrix's spectral interval, never widen it \citep[Thm.~1]{oswald2026expressivity}. Once any ReEig application lands in $[\epsilon,\infty)$, every subsequent layer's input does too. This means that at any stacking depth, the whole BiMap$\to$ReEig stack collapses to two BiMap layers around a single nonlinear activation, whether or not that activation ever clips.

In practice, the resulting structure is even less expressive than this collapsed block: ReEig not activating on real EEG data has been observed in the literature \citep{kobler2022spd,wilson2025deepriemannian}, and in our
own datasets (Appendix \ref{app:reeig-activation}). This means the network's manifold-valued layers often collapse not merely to two BiMap operations around one activation, but to a single fixed linear transform, regardless of depth.

\begin{figure}[h]
\centering
\includegraphics[width=0.6\linewidth]{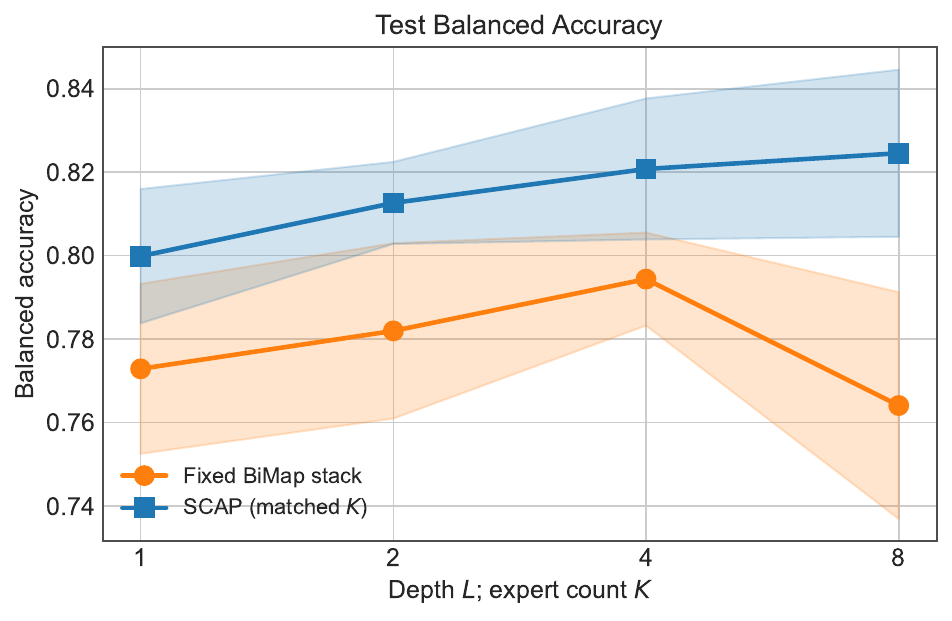}
\caption{Real-EEG validation of depth-closure, Weibo2014 ($n{=}60$,
         $k{=}30$, $m{=}9$ domains; 5-fold CV, mean $\pm$ 1 std). Test balanced accuracy, fixed BiMap$\to$ReEig
         depth $L$ vs.\ SCAP with $K$ matched to $L$.}
\label{fig:depth-empirical}
\end{figure}

We verified this on Weibo2014 \citep{weibo2014db} across the full depth
sweep ($L\in\{1,2,4,8,9\}$), and at $L{=}1$ on two datasets, BNCI2015001 and BNCI2014001: ReEig never activates on any of the three, with every eigenvalue at least $9\times$ above its clamp threshold $\epsilon$; \Cref{app:reeig-activation} gives the full distribution and margins. We
verify the consequence in stacked SPDNet's accuracy against SCAP trained
with $K$ experts matched to each $L$ (\Cref{fig:depth-empirical}). In all
comparison tests through different depths, SPDNet accuracy is statistically indistinguishable, including between the shallowest and deepest stack ($n{=}5$ folds, Holm-corrected), illustrating exactly what closure predicts: without an active ReEig, no larger function class is reachable, so no depth should reliably beat another. Differently, SCAP sits above the depth stack at every point while using no more filters than stacked SPDNet, with a consistent, large effect across all four comparisons (+2.6 to +6.1 points), though, like the depth-only result, none individually passes Holm correction, likely due to the low fold count.

\section{Stiefel Cross-Attention Pool (SCAP)}
\label{sec:scap}

The proposed SCAP layer can be described by a strictly broader filter family than that of BiMap:

\begin{definition}[Routed family]
\label{def:routed-family}
\[
\mathcal{F}_k^{\mathrm{SCAP}} = \Bigl\{\, X \mapsto W(X)^\top X\, W(X) \;\Bigm|\; W(X) = \mathcal{R}_{W_\text{base}}\Bigl(\sum_{j=1}^K \alpha_j(X)\,\Pi_{W_\text{base}}(W_j)\Bigr),\ \alpha(X) \in \Delta^{K-1} \Bigr\}
\]
for a fixed anchor $W_\text{base} \in \St(n,k)$ and component filters
$W_1,\ldots,W_K \in \St(n,k)$, with $\Pi_{W_\text{base}}$,
$\mathcal{R}_{W_\text{base}}$ as in \Cref{def:tangent-retraction}.
\end{definition}

Taking $\alpha(X)$ constant collapses $\mathcal{F}_k^{\mathrm{SCAP}}$ back to
$\mathcal{F}_k$, so the family's extra capacity comes entirely from the input-dependence of the combination of experts.

SCAP's per-sample routing combines its $K$ experts' tangent images at $W_\text{base}$, so it can reach any point in their convex hull, not just the $K$ experts themselves. Matching $m$ domains' dedicated filters therefore does not require $K=m$: if their optimal filters vary along only $r$ directions near $W_\text{base}$, $K=r+1$ experts suffice, with no saving in the fully-separated case of \Cref{sec:pigeonhole}, where generically $r=m-1$. This guarantee is first-order: its approximation error grows with the distance between the domains' optimal filters and $W_\text{base}$ (\Cref{prop:expert-compression}). SCAP thus acts, to first order, as a compressor of the filter bank, not a discrete lookup table indexed by domain.

In order to make the retraction step in \Cref{def:routed-family} unbiased, we define $W_\text{base}$ as a dedicated, freely-learned anchor detached from the expert pool, since using one of the experts as basepoint would privilege it by inducing less tangent projection distortion regardless of its merit (Fix 1). The input-dependent weights $\alpha(X)$ used for combining tangent expert values are obtained by cross-attending input-dependent queries $q_i$ to the keys associated with the pool's values, as in the L2P \citep{wang2022learning} framework. Differently from their solution, SCAP's queries are obtained by projecting the input in a basis maximizing the inter-domain variance, presenting a lighter alternative to the Vision Transformer \citep{dosovitskiy2020vit} features used in L2P (Fix 2). In both mechanisms, query $q_i$ and key $\mathcal{E}_j$ vectors are compared in ambient Euclidean space. Differently from the mechanism presented in \citep{wang2022learning}, in which values are combined by mere equal-importance concatenation, SCAP combines manifold experts through weighted geometry-preserving operations as specified in \Cref{def:routed-family}. Finally, in order to align the trainable keys $\mathcal{E}_j$ to the query $q_i$ without collapsing the pool toward a shared mean, we employ a winner-take-all alignment loss, as in sparsely-gated mixture-of-experts routing \citep{shazeer2017outrageously,fedus2022switch} (Fix 3). Ablations showing the effect of Fixes 1--3 can be seen in \Cref{app:scap-fixes}.

\subsection{The Full Layer}
\label{sec:full-layer}
Given a sample $X_i$ with domain identity $d_i$, the tangent representation at the identity is
\[
  x_i = \upper(\log X_i) \in \R^{n(n+1)/2},
\]
the matrix logarithm's upper triangle with $\sqrt2$-scaled off-diagonal entries \citep{pennec2006}. Above a threshold $\rho^\ast=50$ on the tangent-to-domain ratio $\rho = n(n+1)/(2m)$ — which also governs Fix 3 (\Cref{app:scap-fixes}) — $x_i$ is first routed through a frozen, domain-separating projection $P_\text{sep}$ (Fix 2, full construction in \Cref{app:scap-fixes}):
\[
  \tilde x_i = \begin{cases} x_i P_\text{sep} & \rho > \rho^\ast \\ x_i & \rho \le \rho^\ast \end{cases}.
\]
The query concatenates this representation with a learned domain embedding,
\[
  q_i = \mathrm{MLP}\bigl([\tilde x_i;\ \mathrm{Emb}(d_i)]\bigr) \in \R^{d_\text{att}},
\]
and attends to the $K$ expert keys stacked in $\mathcal{E} \in \R^{K\times d_\text{att}}$:
\[
  \alpha_i = \softmax\!\left(\frac{q_i \mathcal{E}^\top}{\sqrt{d_\text{att}}}\right) \in \Delta^{K-1}.
\]
The per-sample filter $W_i$ is \Cref{def:routed-family}'s $W(X)$ evaluated at $X_i$, with weights $\alpha_{i,j}$; its BiMap output
\[
  Y_i = W_i^\top X_i W_i \in \SPD(k)
\]
feeds the standard ReEig$\to$LogEig$\to$Linear tail.

\section{Routing Diagnostics}
\label{sec:routing-diagnostics}

If SCAP's routing weights is input-independant, the layer degenerates to one fixed filter and \Cref{sec:pigeonhole}'s ceiling applies ($\alpha(X)$ constant in \Cref{def:routed-family} applied in \Cref{lem:pigeonhole}):

\begin{corollary}[Collapse implies the fixed-filter ceiling]
\label{cor:collapse-ceiling}
If there exists $\bar\alpha \in \Delta^{K-1}$ such that $\alpha(X) \equiv \bar\alpha$ for every $X$, SCAP degenerates to a single fixed filter $f_{\bar W} \in \mathcal{F}_k$, with $\bar W = \mathcal{R}_{W_\text{base}}\bigl(\sum_{j=1}^K \bar\alpha_j\,\Pi_{W_\text{base}}(W_j)\bigr)$ as in \Cref{def:routed-family}, and by \Cref{lem:pigeonhole}, $\frac{1}{m}\sum_d s_d(\bar W) \le k/m$. 
\end{corollary}

SCAP's collapse can nullify its advantage over fixed-filter baselines, even if in some cases classification accuracy can still be higher (\Cref{tab:scap-ablation}) due to different optimization paths.

\paragraph{Why collapse happens.} Gradients
reach $\alpha_i$ only through how far each expert's tangent vector $\Pi_{W_\text{base}}(W_j)$ sits from the current mixture $\bar\xi_i = \sum_j \alpha_{i,j}\Pi_{W_\text{base}}(W_j)$; when experts are similar, these deviations are small and $\alpha_i$ receives negligible gradient regardless of how useful better routing would be. Experts, in turn, stay similar precisely because near-uniform routing gives every expert nearly identical training signal, leaving none with pressure to specialize away from the others. Each side of this loop maintains the condition the other needs to stay stuck: collapse is the default outcome of training the naive layer end-to-end, not a rare failure mode.

\paragraph{Empirical evidence.} The ablation table in \Cref{app:scap-fixes} shows that the naive SCAP layer (before any of \Cref{sec:scap}'s three fixes) collapses exactly this way on Weibo2014: its balanced accuracy matches almost exactly that of a single fixed filter set to the uniform-weight barycenter of the same $K$ experts ($\gap_{K=1} := \bacc_\text{adaptive}-\bacc_{K=1}\approx0$), confirming collapse under \Cref{cor:collapse-ceiling}. Yet both sit $4.5$ points above the plain fixed-BiMap baseline, which is not a contradiction: the Corollary bounds every fixed filter's \emph{capacity}, not its accuracy. This reveals that even within the collapsed regime, jointly training $K$ experts and retracting their barycenter can still reach a better point within that same capped class $\mathcal{F}_k$ than the single optimization path ordinary BiMap training follows from one initialization. The gain is real but comes from optimization, not from routing. Applying the three fixes was observed to increase domain-routing alignment and decrease routing entropy relative to this naive layer, detailed alongside the full metric definitions in \Cref{app:scap-fixes}.

\section{Experiments}
\label{sec:experiments}

We evaluate the effect of replacing SPDNet's BiMap layer with SCAP on five EEG motor-imagery datasets: Weibo2014
\citep{weibo2014db}, BNCI2015001 \citep{faller2012autocalib},
BNCI2014001 \citep{tangermann2012review}, GrosseWentrup2009
\citep{grossewentrup2009beamforming}, and Schirrmeister2017
\citep{schirrmeister2017deep}. Every dataset is evaluated under binary
motor-imagery classification (left vs.\ right hand, or hand vs.\ foot, depending on the dataset). Domains are subject-session combinations, following cross-domain protocol: every domain appearing in a test fold is also present, under other trials, in that fold's training set. RPA, BN, and DSBN are the closest reference points for asking whether the added expressivity of \Cref{sec:expressivity-gap} has substantial improvement, not because SCAP is a direct domain-adaptation competitor, given these solutions could even be composed with it, but because they are standard solutions for cross-domain problem. SPDNet, BN and DSBN used where the original implementation (including their BiMap layer) from $\texttt{spd\_learn}$ \citep{aristimunha2026spdlearn} library; SCAP uses the ReEig$\to$LogEig tail from the same repository; RPA (recenter + rotate) uses \texttt{pyRiemann}, followed by tangent-space logistic regression.

\subsection{Main results}
\label{sec:main-results}

\begin{table}[h]
\centering
\caption{SCAP vs.\ the plain fixed-BiMap baseline. Balanced accuracy,
5-fold cross-domain evaluation. $p$: raw paired-test $p$-value.}
\label{tab:baseline-results}
\begin{tabular}{lcc}
\toprule
Dataset & Baseline (SPDNet with BiMap) & SCAP (SPDNet with SCAP) \\
\midrule
Weibo2014 & .773 & \textbf{.825} ($p{=}.0098$) \\
GrosseWentrup2009 & .678 & \textbf{.726} ($p{=}.0158$) \\
BNCI2015001 & .757 & \textbf{.807} ($p{=}.0041$) \\
BNCI2014001 & .801 & \textbf{.839} ($p{=}.0135$) \\
Schirrmeister2017 & .865 & \textbf{.918} ($p{=}.0156$) \\
\bottomrule
\end{tabular}
\end{table}

Against the plain fixed-BiMap baseline, which is the comparison the paper's theory directly targets, SCAP outperforms it on all 5 datasets. Within that 5-dataset family, all 5 comparisons remain significant after Holm correction (raw $p\in[.004,.016]$, Holm-corrected $p\in[.021,.041]$).

\begin{table}[h]
\centering
\caption{SCAP vs.\ domain-adaptive baselines. Balanced accuracy, 5-fold
cross-domain evaluation. Bold: best per dataset. $p$: raw paired-test
$p$-value for SCAP vs.\ that method.}
\label{tab:main-results}
\begin{tabular}{lcccc}
\toprule
Dataset & SCAP & RPA & BN & DSBN \\
\midrule
Weibo2014 & \textbf{.825} & .720 ($p{=}.002$) & .783 ($p{=}.006$) & .802 ($p{=}.328$) \\
GrosseWentrup2009 & \textbf{.726} & .700 ($p{=}.368$) & .676 ($p{=}.055$) & .675 ($p{=}.027$) \\
BNCI2015001 & \textbf{.807} & .783 ($p{=}.063$) & .758 ($p{=}.0002$) & .772 ($p{=}.006$) \\
BNCI2014001 & \textbf{.839} & .831 ($p{=}.245$) & .802 ($p{=}.0001$) & .808 ($p{=}.022$) \\
Schirrmeister2017 & .918 & \textbf{.960} ($p{=}.006$) & .855 ($p{=}.002$) & .851 ($p{=}.001$) \\
\bottomrule
\end{tabular}
\end{table}

Significance is assessed per comparison type, using a separate
5-dataset Holm family for each of SCAP-vs-RPA, SCAP-vs-BN, and
SCAP-vs-DSBN pairings. SCAP has the highest
point estimate on 4 of these 5 datasets, with the sole exception of
Schirrmeister2017, where RPA leads. Against BN, SCAP wins on all 5 datasets in point estimate, holding up under Holm correction on 4 (all but GrosseWentrup2009, raw $p{=}.0547$). Against DSBN, SCAP wins on all 5 in point estimate and 2 using Holm correction (BNCI2015001, Schirrmeister2017); under the less conservative BH-FDR criterion, two further datasets also hold up (BNCI2014001, GrosseWentrup2009). Against RPA, 2 of 5 comparisons are conclusive under either correction: Weibo2014 favoring SCAP, Schirrmeister2017 favoring RPA. Routing is strongly domain-structured on all five datasets (\Cref{tab:routing-diagnostics}), and a query-ablation control confirms domain identity as the dominant routing signal, with input-conditioning a smaller but consistent contributor (\Cref{app:query-ablation}).

\section{Conclusion}
\label{sec:conclusion}
SCAP targets bilinear congruence filters on SPD data wherever a single filter layer must serve multiple, competing domains at once. Its expressivity fix also reads as domain-adaptive, acting on different manifold transforms than deep adaptive baselines such as BN and DSBN (\Cref{sec:related-work}). The capacity ceiling proved for the worst case of mutually orthogonal domain subspaces (\Cref{sec:pigeonhole}) constrains any single fixed filter as BiMap; \citet{oswald2026expressivity} shows a whole SPD network reduces architecturally to one BiMap$\to$ReEig$\to$BiMap block regardless of intended depth. We show this reduces further, to a single fixed filter, on the EEG datasets where ReEig never activates (\Cref{sec:depth-closure}, \Cref{app:reeig-activation}), constraining those networks to the capacity ceiling. SCAP overtakes this ceiling as domains grow, and matches a dedicated filter per domain to first order with fewer experts than domains when their optimal filters vary along few directions near a tangent-space basepoint (\Cref{prop:expert-compression}). On real data, it significantly improves over standard SPDNet with BiMap on all five EEG datasets tested (\Cref{sec:experiments}). The same architectural collapse could plausibly extend to medical-imaging and radar applications, wherever SPD matrices are well-conditioned enough that ReEig doesn't activate. Extending that demonstration to multi-class and cross-paradigm decoding (e.g., P300, SSVEP), to those medical-imaging and radar settings, and sweeping $K$ against \Cref{app:expert-compression}'s compression bound rather than fixing it by rule, are natural next steps.
\newpage


\subsection*{AI use statement}
In this work, we used generative AI tools (large language models, including an agentic coding assistant) to implement methods, i.e.\ to write and refactor parts of the training, evaluation, and statistical-testing code under the authors' specification, to implement the toy dataset in  \Cref{app:toy-experiment}; to provide feedback on research methodology and experimental design; and to assist in refining the statements of mathematical claims and the writing of their proofs, (\Cref{lem:pigeonhole}, \Cref{prop:expert-compression}, \Cref{app:expert-compression});  all the interventions were revised by the authors and corrected where judged necessary. We have not used generative AI tools for the solution conception, neither to generate or preprocess real datasets: all EEG data are public recordings treated in our own code or using public libraries. Additionally, we used generative AI tools to edit the paper for readability, to draft parts of sections, to create or edit figure code, and to format references.
AI-written text and code were reviewed by the authors, and all reported numbers were produced by code the authors ran and checked. We take responsibility for the final content of this work, including text, claims, or artifacts produced with the aid of generative AI.

\subsection*{Ethics statement}
This work uses only previously published, publicly available EEG motor-imagery recordings (Weibo2014, BNCI2015001, BNCI2014001, GrosseWentrup2009, Schirrmeister2017), accessed through the FII BCI Corpus \citep{fii-corpus-mi}. They were acquired by their original authors under the ethics and informed-consent procedures reported in the original publications, and are distributed in de-identified form. No new data were collected, and no attempt was made to re-identify participants; subject exclusions were based on signal quality only (\Cref{app:experimental-details}). SCAP conditions on subject-session identity and requires labelled trials from every domain it serves; it is intended for assistive and rehabilitation BCIs and was neither designed nor evaluated for identifying individuals from their EEG. The authors declare no conflicts of interest.

\subsection*{Reproducibility statement}
All assumptions and complete proofs are given in \Cref{sec:pigeonhole}
(\Cref{lem:pigeonhole}), \Cref{app:tangent-retraction} (the projection and
retraction used throughout), and \Cref{app:expert-compression} (\Cref{prop:expert-compression}). The synthetic experiment is fully specified in \Cref{app:toy-experiment}. For the EEG experiments, preprocessing, domain
definition, per-dataset $m$ and $K$, optimizer, early stopping, Stiefel
parametrization, seeds, and compute are reported in \Cref{app:experimental-details}; the complete SCAP specification (frozen
projection, thresholds, loss weights) and all routing-metric definitions are in \Cref{sec:full-layer} and \Cref{app:scap-fixes}. All datasets are public
\citep{fii-corpus-mi,chevallier2024moabb}, and baselines use public
implementations (\texttt{spd\_learn} \citep{aristimunha2026spdlearn},
\texttt{pyRiemann}). Anonymized source code for SCAP, the baselines, the
synthetic experiment, and the significance tests is provided as supplementary
material.


\bibliographystyle{plainnat}

\appendix
\section{Core Notation}
\label{app:notation}

\begin{center}
\begin{longtable}{@{}ll@{}}
\caption{Core notation used throughout the paper. Symbols local to a single
derivation (e.g.\ the toy-experiment construction of \Cref{app:toy-experiment})
are defined where used and omitted here.}
\label{tab:notation} \\
\toprule
Symbol & Meaning \\
\midrule
\endfirsthead
\multicolumn{2}{@{}l}{\tablename\ \thetable{} -- continued from previous page} \\
\toprule
Symbol & Meaning \\
\midrule
\endhead
\midrule
\multicolumn{2}{r@{}}{continued on next page} \\
\endfoot
\bottomrule
\endlastfoot
\multicolumn{2}{@{}l}{\textit{Manifolds and core objects}} \\
\midrule
$n$, $k$ & input / output dimension of a BiMap filter \\
$m$ & number of domains \\
$K$ & number of experts (distinct from the filter dimension $k$) \\
$L$ & stacking depth of a BiMap$\to$ReEig network \\
$W \in \St(n,k)$ & a single Stiefel filter \\
$U_d \in \St(n,k)$ & domain $d$'s $k$-dimensional signal subspace \\
$s_d(W)$ & per-domain subspace overlap, $\|U_d^\top W\|_F^2$ \\
$\mathcal{F}_k$, $\mathcal{F}_k^{\mathrm{SCAP}}$ & fixed-filter class; SCAP's routed family \\
$\xi \in T_W\St(n,k)$ & generic tangent vector at basepoint $W$ \\
$\Pi_W$, $\mathcal{R}_W$ & tangent projection, retraction at $W$ (\Cref{def:tangent-retraction}) \\
$\epsilon$ & ReEig eigenvalue threshold \\
\midrule
\multicolumn{2}{@{}l}{\textit{SCAP layer (\Cref{sec:scap}, \Cref{sec:full-layer})}} \\
\midrule
$W_\text{base} \in \St(n,k)$ & learnable interpolation anchor (Fix 1) \\
$W_1,\ldots,W_K \in \St(n,k)$ & the pool's component (expert) filters \\
$\xi_j := \Pi_{W_\text{base}}(W_j)$ & expert $j$'s tangent image at $W_\text{base}$ \\
$\alpha(X) \in \Delta^{K-1}$, $\alpha_i$ & routing distribution; its value on sample $i$ \\
$W(X)$, $W_i$, $Y_i$ & per-input / per-sample filter; SCAP's BiMap output \\
$x_i$, $\tilde x_i$ & tangent rep.\ of $X_i$ at identity; after $P_\text{sep}$ (Fix 2) \\
$\rho$, $\rho^\ast$ & tangent-to-domain ratio $n(n{+}1)/2m$; its threshold ($=50$) \\
$P_\text{sep}$, $r_\text{sep}$ & frozen domain-separating projection (Fix 2); its fixed rank ($=40$) \\
$d_i$, $\mathrm{Emb}(d_i)$ & sample $i$'s domain identity; its learned embedding \\
$q_i$, $\mathcal{E}$, $\mathcal{E}_j$ & query; stacked expert keys; expert $j$'s key \\
$d_\text{att}$ & query/key and domain-embedding dimension ($=20$) \\
\midrule
\multicolumn{2}{@{}l}{\textit{Losses and routing diagnostics}} \\
\midrule
$\mathcal{L}_\text{CE}$, $\mathcal{L}_\text{align}$, $\mathcal{L}$ & classification, alignment, total loss \\
$\lambda_\text{align}$, $j_i^\ast$, $\mathrm{sg}(\cdot)$ & alignment weight; winner-key index; stop-gradient (Fix 3) \\
$\bacc$ & balanced accuracy \\
$\gap_\text{base}$, $\gap_{K=1}$ & bAcc gain over fixed BiMap; over uniform-barycenter $K{=}1$ \\
$\bar H$ & mean routing entropy, $\frac{1}{N\log K}\sum_i H(\alpha_i) \in [0,1]$ \\
$r_d$ & domain-routing alignment (domain-mean vs.\ per-sample routing variance) \\
\midrule
\multicolumn{2}{@{}l}{\textit{Expert compression (\Cref{app:expert-compression})}} \\
\midrule
$W_d^\ast$, $\delta_d$ & domain $d$'s optimal filter; its tangent image at $W_\text{base}$ \\
$r$ & affine dimension of $\{\delta_1,\ldots,\delta_m\}$; governs $K=r{+}1$ \\
\end{longtable}
\end{center}

\section{Tangent Space, Projection, and Retraction on the Stiefel Manifold}
\label{app:tangent-retraction}

The tangent space to $\St(n,k)$ at $W$ is $T_W\St(n,k) = \{\xi \in \R^{n\times k} : W^\top\xi + \xi^\top W = 0\}$, the kernel of the surjective linear map $\xi\mapsto\mathrm{sym}(W^\top\xi)$ onto $\mathrm{Sym}(k)$, and therefore a linear subspace of $\R^{n\times k}$ of dimension $nk-\tfrac{k(k+1)}{2}$ for every $W$ \citep{edelman1998geometry}. This is the quantity \Cref{prop:expert-compression} uses to bound $r$.

\begin{definition}[Tangent projection and retraction]
\label{def:tangent-retraction}
Given a second point $W' \in \St(n,k)$, its first-order logarithm at $W$ is
\[
  \Pi_W(W') = W' - W\,\mathrm{sym}(W^\top W'), \qquad
  \mathrm{sym}(A) = \tfrac{1}{2}(A+A^\top),
\]
the orthogonal projection of $W'$ onto $T_W\St(n,k)$. Conversely, given $\xi \in T_W\St(n,k)$, the sign-corrected QR retraction
\[
  \mathcal{R}_W(\xi) = Q\,\mathrm{diag}(\mathrm{sign}(\mathrm{diag}\,R)),
  \qquad W+\xi = QR \ \text{(thin QR decomposition)},
\]
maps back onto $\St(n,k)$, satisfying $\mathcal{R}_W(0)=W$ and $D\mathcal{R}_W(0)=\mathrm{id}$ \citep{absil2008optimization, edelman1998geometry}.
\end{definition}

\Cref{def:routed-family} applies $\Pi_{W_\text{base}}$ to each expert $W_j$ and $\mathcal{R}_{W_\text{base}}$ to their combination.

\section{Toy Experiment: Technical Construction}
\label{app:toy-experiment}

\Cref{fig:overlap-ceiling}'s toy experiment uses synthetic data where each domain's discriminative subspace is known exactly by construction, so trained models can be checked against a closed-form prediction rather than only against each other. 

\paragraph{Ambient space and domain subspaces.} Ambient dimension $n{=}24$, filter/subspace dimension $k{=}2$. For $m$ domains ($m \in \{2,3,4,6,8,12\}$ across separate runs, each requiring $mk\le n$), domain $d$ is assigned a private, non-overlapping pair of coordinate axes, 
\[
  U_1 = \mathrm{span}(e_1,e_2), \quad U_2 = \mathrm{span}(e_3,e_4), \quad
  \ldots, \quad U_d = \mathrm{span}(e_{2d-1},e_{2d}).
\]
For $m{=}3$, e.g., $U_1,U_2,U_3$ use coordinates $\{1,2\},\{3,4\},\{5,6\}$; coordinates $7$--$24$ belong to no domain (pure noise for every domain). Being disjoint coordinate axes, the $U_d$ satisfy \Cref{sec:pigeonhole}'s mutual-orthogonality assumption automatically, with no orthogonalization step needed. Let $P_{U_d}\in\R^{24\times24}$ be the diagonal $0/1$ projector onto $U_d$ (e.g.\ $P_{U_1}=\mathrm{diag}(1,1,0,\ldots,0)$). 

\paragraph{Class covariances and the separation parameter.} For each domain $d$, the two class centers are
\[
  \Sigma_{d,+} = \exp(\gamma P_{U_d}) = I_{24} + (e^\gamma-1)P_{U_d},
  \qquad
  \Sigma_{d,-} = \exp(-\gamma P_{U_d}) = I_{24} - (1-e^{-\gamma})P_{U_d},
\]
using $P_{U_d}^2=P_{U_d}$ to collapse the matrix exponential to closed form. Both are $24\times24$ SPD matrices: eigenvalue $e^{\pm\gamma}$ (mult.\ 2, on $U_d$'s two coordinates), eigenvalue $1$ elsewhere (mult.\ 22, no signal). $\gamma{>}0$ controls how far apart the two classes sit on domain $d$'s own coordinates, i.e.\ how easy the classification problem is; the figure uses $\gamma{=}1.2$ ($e^{1.2}{\approx}3.32$, $e^{-1.2}{\approx}0.30$).  

\paragraph{Sampling procedure.} Class centers are never observed directly.  Each trial is a noisy draw,
\[
  X = \frac{1}{\nu}\,ZZ^\top, \qquad
  \text{columns of } Z \sim \mathcal{N}(0,\,\Sigma_{d,\text{class}}),
  \qquad \nu = n+15 = 39,
\]
a Wishart-type draw with $\nu$ degrees of freedom, mimicking finite-trial covariance estimation noise. $\nu{=}n{+}15$, rather than the minimum $\nu{=}n$ required for almost-sure invertibility, gives a moderate, realistic noise level. 40 trials are drawn per (domain, class), split 70/30 train/test, stratified by domain and class. 

In SCAP construction, we use $K=m$ experts (the number \Cref{prop:expert-compression} requires when domains share no structure ($r=m-1$)), and we report $\frac1m\sum_d \E_{X\in d}\,[s_d(W(X))]$ as average overlap for test samples, since the filter is input-specific.


\section{Experimental Details}
\label{app:experimental-details}
Here we report the experimental details of \Cref{sec:experiments} for
reproducibility purposes.

\paragraph{Preprocessing.} Raw EEG trials are band-pass filtered at
8-32\,Hz, rejected under an artifact criterion of \texttt{upperLimit}=1.2, and encoded into sample covariance matrices (SCM) via \texttt{Eegle.jl}'s Riemannian encoding \citep{congedo2026eeglejl}, drawing from the FII BCI Corpus\citep{fii-corpus-mi}, which stores each recording at  $\le256$\,Hz (200\,Hz for Weibo2014, 256\,Hz for BNCI2015001, and 250\,Hz for BNCI2014001, GrosseWentrup2009, and Schirrmeister2017); no further resampling is applied here. One subject was excluded from each of Weibo2014 and Schirrmeister2017 for poor signal quality; \Cref{tab:domain-counts} reports the resulting subject and domain counts, and the class pair used, for all five datasets. Spatial covariance matrices are then preconditioned by per-domain whitening with variance equalization and by recentering each domain's distribution toward $I_n$, via Tikhonov-regularized shrinkage as described in \citep{congedo2026pseudoaffine}. For GrosseWentrup2009 and Schirrmeister2017, this recentering step also reduces dimensionality: $n{=}101$ is a post-reduction dimension (raw channel count 128), obtained by whitening the Euclidean barycenter of each domain's covariance set and truncating to the top eigendirections \citep{congedo2026pseudoaffine}; this triggers automatically above 64 raw channels, and leaves the other three datasets ($n_\text{raw}\le64$) unreduced. 

\begin{table}[h]
\centering
\caption{Domains and experts per dataset. Domain = subject-session, as
stated in \Cref{sec:experiments}. Nominal subject counts per MOABB
\citep{chevallier2024moabb}; $m$ excludes subjects dropped for poor signal quality.}
\label{tab:domain-counts}
\begin{tabular}{lccccc}
\toprule
Dataset & Subjects (nominal) & Subjects used & $m$ & Classes & $K$ \\
\midrule
Weibo2014          & 10 & 9  & 9  & feet / right hand  & 8  \\
BNCI2015001        & 12 & 12 & 28 & feet / right hand  & 16 \\
BNCI2014001        & 9  & 9  & 18 & feet / right hand  & 8  \\
GrosseWentrup2009  & 10 & 10 & 10 & left / right hand  & 9  \\
Schirrmeister2017  & 14 & 13 & 13 & left / right hand  & 12 \\
\bottomrule
\end{tabular}
\end{table}

\paragraph{Expert count.} \Cref{tab:domain-counts} reports the expert
count $K$ used for each dataset alongside its domain count $m$ and classpair. For Weibo2014, GrosseWentrup2009, and Schirrmeister2017,  $K=m-1$. For BNCI2014001 and BNCI2015001, $K$ was fixed early in development ($K{=}8$ and $K{=}16$ respectively) and does not follow this convention, including the specific subject-vs.\ subject-session domain granularity used to define $m$ elsewhere in this paper; we report results as originally run rather than retroactively rerunning with a corrected $K$, since our results, other than optimization effects, should not be sensitive to this choice by theory: SCAP's routed family only requires $K \geq r{+}1$ experts to match a dedicated per-domain filter bank to first order (\Cref{prop:expert-compression}), so an under-provisioned $K$ can only ever cost potential capacity, not create an unfair advantage, whichever granularity was actually used. We do not sweep $K$ against accuracy for the results reported here. 

\paragraph{Training.} Adam \citep{kingma2014adam}, learning rate 0.01, batch size 32, up to 40 epochs with early stopping (patience 10 on validation balanced accuracy, best checkpoint restored), no weight decay. The anchor $W_\text{base}$ and the $K$ expert filters $W_1,\ldots,W_K$ are each maintained on $\St(n,k)$ throughout training via PyTorch's built-in orthogonal parametrization (\texttt{torch.nn.utils.parametrizations.orthogonal}), which reparametrizes each as a differentiable function of an unconstrained matrix so that standard Adam updates preserve orthonormality automatically. This is separate from the per-sample retraction $\mathcal{R}_{W_\text{base}}$ of \Cref{def:tangent-retraction}: at each forward pass, expert tangent images are computed via a first-order Riemannian log at $W_\text{base}$, averaged with the routing weights $\alpha(X)$, and mapped back to $\St(n,k)$ via an explicit sign-corrected QR retraction, matching \Cref{def:tangent-retraction} exactly. A single fixed seed per run; results in \Cref{tab:baseline-results,tab:main-results} are averaged over the 5 cross-validation folds under this seed.

\paragraph{Compute.} 12-core CPU node, 12 threads pinned for run-to-run determinism. Baseline BiMap averages 55\,s/fold  (23\,min over all five datasets); SCAP averages 2.4\,min/fold (61\,min total), a $\sim2.6\times$ overhead from the per-sample Stiefel log/exp retraction and the frozen-projection precompute. 

\begin{table}[h]
\centering
\caption{Trainable parameters.}
\begin{tabular}{lccc}
\toprule
Dataset & Baseline (BiMap) & SCAP & Ratio \\
\midrule
Weibo2014          & 2{,}732 & 20{,}732 & 7.6$\times$ \\
BNCI2015001        &   314   &  8{,}990 & 28.6$\times$ \\
BNCI2014001        &   862   & 16{,}682 & 19.4$\times$ \\
GrosseWentrup2009  & 7{,}602 & 56{,}692 & 7.5$\times$ \\
Schirrmeister2017  & 7{,}602 & 71{,}962 & 9.5$\times$ \\
\bottomrule
\end{tabular}
\end{table}

\section{ReEig Activation on Real EEG Data}
\label{app:reeig-activation}

\begin{figure}[h]
\centering
\includegraphics[width=\linewidth]{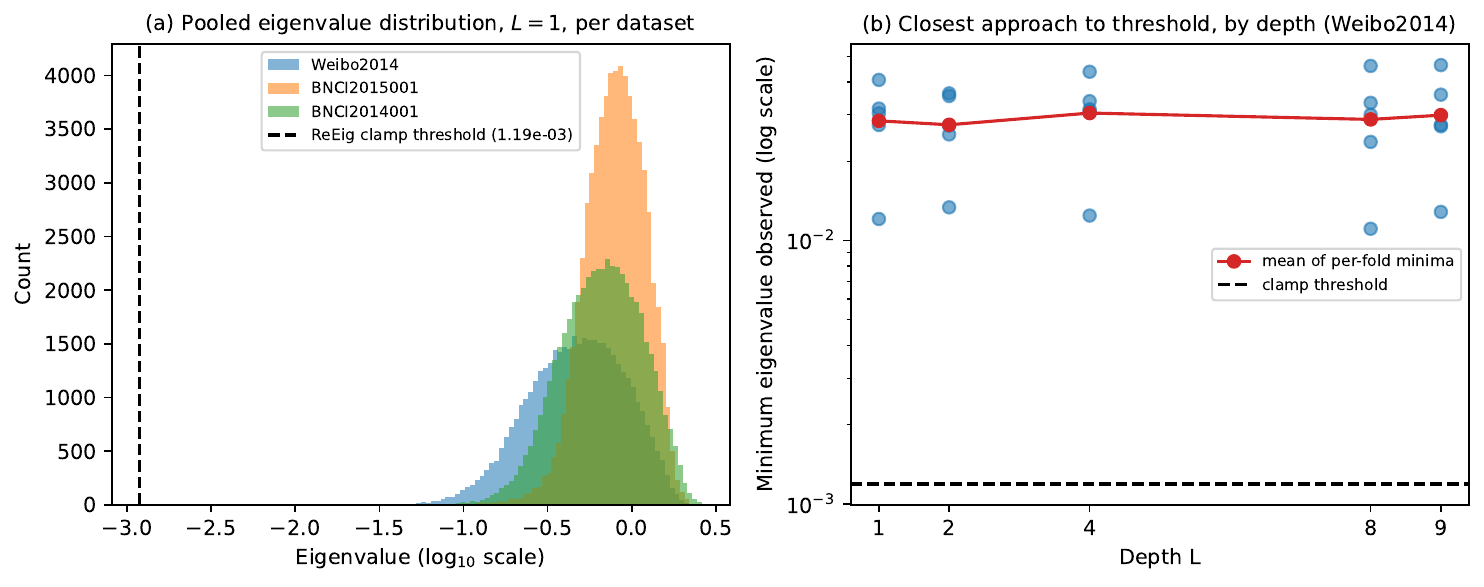}
\caption{ReEig eigenvalue activation vs.\ clamp threshold
$\epsilon\approx1.19\times10^{-3}$ (dashed). \textbf{(a)} Pooled
$L{=}1$ eigenvalue distribution, three real EEG datasets — margin
over $\epsilon$: Weibo2014 $10.1\times$, BNCI2015001 $10.0\times$,
BNCI2014001 $11.1\times$. \textbf{(b)} Minimum eigenvalue by depth
$L\in\{1,2,4,8,9\}$, Weibo2014 only.}
\label{fig:reeig-activation}
\end{figure}

\Cref{sec:depth-closure} reports that ReEig never activates on Weibo2014 across every depth, layer, and fold tested (1{,}013{,}040 eigenvalues checked; global minimum margin $9.3\times$ over $\epsilon$, pooled across $L\in\{1,2,4,8,9\}$). We additionally verified first-layer ($L{=}1$) activation on two further real EEG datasets, BNCI2015001 and BNCI2014001, following the same protocol; all three datasets clear the clamp threshold by at least $10\times$ at $L{=}1$ (\Cref{fig:reeig-activation}), with no non-positive or anomalous eigenvalues observed. Panel (b) additionally tests, for Weibo2014 (the only dataset for which we trained a full depth sweep) whether deeper stacks drift closer to activating the nonlinearity even without crossing it; they do not, the mean minimum eigenvalue per fold stays flat from $L{=}1$ to $L{=}9$, ruling out a depth-dependent proximity effect as an explanation for the accuracy degradation observed at high $L$. We have not yet verified this behavior on GrosseWentrup2009 or Schirrmeister2017.
\section{Expert Compression: Proof}
\label{app:expert-compression}

Each domain $d$ has an ideal filter $W_d^\ast$, the BiMap projection that would perform best if trained exclusively on that domain's own data, as in a fully dedicated per-subject calibration. SCAP does not give each domain a private filter; it keeps $K$ shared experts and blends them per sample around a learned anchor. \Cref{prop:expert-compression} answers: how many experts are needed to guarantee every domain's ideal filter is reachable as some blend of the $K$ experts?

We call this quantity $r$, the number of independent directions of variation across all $m$ domains' ideal filters, measured in the tangent space at the shared anchor. If what differs between subjects and sessions is driven by a small number of common factors as
electrode-cap placement drift, a between-session recording-condition shift, broadly shared demographic or anatomical variation, $r$ can stay small even as $m$ grows. If every domain varies in a way unrelated to every other domain, $r$ approaches $m-1$ and no compression is available (\Cref{sec:pigeonhole}'s worst-case construction is exactly this regime).

\begin{proposition}[Expert compression]
\label{prop:expert-compression}
Let $W_\text{base} \in \St(n,k)$ be SCAP's interpolation basepoint (\Cref{def:routed-family}). For each domain $d \in \{1,\ldots,m\}$, let $W_d^\ast \in \St(n,k)$ be the maximizer of $s_d(\cdot)$ (\Cref{def:overlap}) closest to $W_\text{base}$ in Frobenius norm, and let $\delta_d = \Pi_{W_\text{base}}(W_d^\ast)$ (\Cref{def:tangent-retraction}) be its tangent image. Define
\[
  r := \dim\,\mathrm{aff}(\{\delta_1,\ldots,\delta_m\}) \;\le\;
  \min\bigl(m-1,\ \dim T_{W_\text{base}}\St(n,k)\bigr),
\]
the affine dimension of the domain-optimal tangent directions. 

There exist $K = r+1$ expert filters and, for every domain $d$, convex weights $\alpha^{(d)} \in \Delta^{K-1}$,

such that SCAP's interpolated filter for domain $d$ agrees with $W_d^\ast$ to first order, i.e.\ up to an error $o(\|W_d^\ast-W_\text{base}\|_F)$, provided some $r$-simplex enclosing $\{\delta_1,\ldots,\delta_m\}$ has its vertices in $\Pi_{W_\text{base}}(\St(n,k))$. Every domain's overlap $s_d(W_d^\ast)$ is thus matched to first order simultaneously, using $K = r+1 \le m$ stored expert matrices rather than the $m$ an explicit one-filter-per-domain bank requires. 
\end{proposition}

This hypothesis makes the guarantee local. In the proof, the experts' tangent images are the corners of a simplex that encloses the domain targets $\delta_d$, so that each target is reachable as a convex combination of experts. Since an expert's tangent image can lie only a bounded distance from the anchor, these corners must remain within that reach. This holds whenever the domains' optimal filters are close enough to $W_\text{base}$, which is also where the first-order approximation is accurate. Without compression ($r=m-1$) it holds automatically, since the targets $\delta_d$, each the tangent image of $W_d^\ast$, then form the simplex themselves; it can fail only when compression is sought ($m>r+1$) and the targets are spread out, so that the enclosing simplex must extend beyond them. In the mutually orthogonal worst case of \Cref{sec:pigeonhole}, $r=m-1$ generically, so the hypothesis holds and it is the first-order accuracy that degrades.

\begin{proof}
\textbf{Necessity.} Every point SCAP's router can produce is a convex combination of the $K$ experts' tangent images (its weights lie in $\Delta^{K-1}$), and every convex combination is in particular an affine combination, so the reachable set lies within the experts' affine span, itself of dimension at most $K-1$. If that span contains all $m$ targets $\delta_d$, then $r \le K-1$, i.e.\ $K \ge r+1$.

\textbf{Sufficiency.} By hypothesis, each vertex is the tangent image $\Pi_{W_\text{base}}(W_j)$ of some expert $W_j \in \St(n,k)$; setting the $K=r+1$ experts accordingly, every $\delta_d$ is a convex combination of their tangent images, which is exactly what SCAP's softmax-simplex router expresses.
\end{proof}

This result holds in the tangent-space linearization at $W_\text{base}$, i.e.\ only to first order: the approximation error grows with $\|W_d^\ast - W_\text{base}\|_F$. The guarantee is therefore informative when domains' optimal filters cluster around a common tangent-space basepoint and weakens as they spread apart. The mutually orthogonal case of \Cref{sec:pigeonhole} is the extreme of this spread: when the domains' subspaces are mutually orthogonal, no single anchor can lie close to all of them, so each domain's target $\delta_d$ sits far from $W_\text{base}$ and the first-order approximation error is largest. This does not limit SCAP's capacity: SCAP serves each domain with its own routed filter, displaced from the anchor along $\delta_d$, so the fixed-filter ceiling of \Cref{lem:pigeonhole} does not apply to it. Additionally, the bound on $r$ is stated against $\dim T_{W_\text{base}}\St(n,k)$ rather than $\dim\St(n,k)$: the two coincide numerically, but it is the tangent space, not the manifold itself, in which the $\delta_d$ and the covering simplex are actually constructed. Separately, $s_d(U_dV)=s_d(U_d)$ for every $V\in O(k)$ (a $k\times k$ change of basis for $U_d$'s subspace, i.e.\ $O(k)=\St(k,k)$), so the maximizer of $s_d(\cdot)$ is unique only up to this rotation; taking $W_d^\ast$ closest to $W_\text{base}$, as specified above, resolves the ambiguity and pins $\delta_d$ down uniquely. This result is a capacity/existence statement, not something the experiments in this paper measure directly.

\section{SCAP's Three Fixes: Full Specification}
\label{app:scap-fixes}

This appendix gives the full construction behind \Cref{sec:scap}'s three
fixes, condensed to one paragraph in the main text.

\paragraph{Fix 1: a dedicated anchor.} \Cref{def:tangent-retraction}'s pattern needs a basepoint to project onto and retract from; the routed family itself leaves that basepoint unspecified. Attaching it dynamically to whichever filter the mixture already favors creates a self-reinforcing bias: an expert close to the current basepoint suffers less tangent-projection distortion than one far from it, and so accumulates further weight regardless of whether it is actually the better filter for the sample at hand. SCAP
breaks this by making $W_\text{base} \in \St(n,k)$ a dedicated learnable parameter, initialized independently of the expert pool and updated freely by the optimizer via an orthogonal reparametrization that keeps it on $\St(n,k)$.

\paragraph{Fix 2: a domain-separating query.} 
The query network is the MLP that maps each sample to its routing query $q_i$ (\Cref{sec:full-layer}). Trained through the classification loss alone, it has no incentive to encode which domain a sample comes from: it learns whatever helps classification, so samples from different domains can produce similar queries and be routed alike. Domain information is present in the input, but hard to reach: the between-domain signal in the tangent vector $x_i$ spans a subspace of rank at most $m-1$ inside a space of dimension $n(n+1)/2$. When that space is large relative to $m$, the domain signal is a small fraction of the query's input, and a task-driven query does not recover it unaided. Fix 2 hands this subspace to the query directly, by projecting $x_i$ onto the directions of largest between-domain variance before it enters the MLP. On Weibo2014, this raises domain-routing alignment $r_d$ from .335 to .690 (\Cref{tab:scap-ablation}).

Write the tangent representation of $X_i$ at the identity as $x_i = \upper(\log X_i) \in \R^{n(n+1)/2}$, the matrix logarithm's upper triangle with $\sqrt2$-scaled off-diagonal entries \citep{pennec2006}. Whether $x_i$ needs help depends on the \emph{tangent-to-domain ratio} $\rho = n(n+1)/(2m)$: a compact tangent space relative to $m$ ($\rho$ small) leaves domain structure directly accessible, while a large ambient tangent space ($\rho$ large) buries it. Above a threshold
$\rho^\ast=50$, we route $x_i$ through a frozen projection
\[
  P_\text{sep} \in \R^{n(n+1)/2 \times r_\text{sep}}, \quad r_\text{sep}{=}40, 
\]
\[
  \text{columns} = \text{top-}r_\text{sep}\text{ eigenvectors of }
  S_B = \textstyle\sum_{d=1}^m n_d(\mu_d-\mu)(\mu_d-\mu)^\top,
\]
where $n_d$, $\mu_d$ are domain $d$'s sample count and mean tangent vector and $\mu$ is the overall mean.  $P_\text{sep}$ is computed once from training data and never touched by gradient descent, playing the same role L2P's frozen backbone plays for its prompts, but implemented differently:
\[
  \tilde x_i = \begin{cases} x_i P_\text{sep} & \rho > \rho^\ast \\
  x_i & \rho \le \rho^\ast. \end{cases}
\]

\paragraph{Fix 3: decoupled key alignment.} Even with a stable,
domain-informative query, expert keys trained by cross-entropy alone chase whatever helps classification, which need not align with domain identity. We instead train keys with a dedicated alignment loss,
\[
  \mathcal{L}_\text{align} = \lambda_\text{align}\,
  \E_i\bigl[1 - \cos(\mathrm{sg}(q_i),\, \mathcal{E}_{j_i^\ast})\bigr], \qquad
  j_i^\ast = \argmax_j \alpha_{i,j},
\]
where $\mathrm{sg}(\cdot)$ stops the gradient through the query and only the top-weighted key $\mathcal{E}_{j_i^\ast}$ updates per sample. Pulling every key toward every query would collapse the pool to its mean; the winner-take-all update instead gives each key competitive pressure to specialize toward the domains that actually favor it. In the high-$\rho$ regime (Weibo2014, GrosseWentrup2009, Schirrmeister2017), keys are detached from the classification gradient and trained exclusively by $\mathcal{L}_\text{align}$ ($\lambda_\text{align}=0.05$). In the low-$\rho$ regime (BNCI2015001, BNCI2014001), where domain structure is already recoverable, Fixes 2 and 3 are both disabled ($\lambda_\text{align}=0$) and keys are trained by the classification loss alone. The total objective is $\mathcal{L} =
\mathcal{L}_\text{CE} + \mathcal{L}_\text{align}$.

\paragraph{Ablation: what breaks the degeneracy?} \Cref{tab:scap-ablation}
isolates each fix's contribution on Weibo2014.Two diagnostics accompany balanced accuracy: the $K{=}1$ proxy gap,
\[
  \gap_{K=1} = \bacc_\text{adaptive} - \bacc_{K=1}, \qquad
  \gap_\text{base} = \bacc_\text{adaptive} - \bacc_\text{baseline},
\]
the first comparing against a single fixed filter set to the uniform-weight barycenter of the same $K$ experts (zero under uniform collapse), the second against the plain fixed-BiMap baseline (\Cref{tab:baseline-results}); and mean routing entropy $\bar H = \frac{1}{N\log K}\sum_i H(\alpha_i) \in [0,1]$ (0: one-hot; 1: uniform), how confidently each sample is routed. Domain-routing alignment $r_d = \mathrm{tr}\,\mathrm{Cov}_d(\bar\alpha_d)\,/\,\mathrm{tr}\,\mathrm{Cov}_i(\alpha_i)$, the variance of the domains' mean routing vectors $\bar\alpha_d$ relative to that of individual samples' routing vectors $\alpha_i$, tests whether domains route differently from one another: values near 1 mean routing is determined almost entirely by domain identity, values near 0 that routing ignores it. Both variances divide by one less than their count, $m-1$ for the $m$ equally weighted domain means and $N-1$ for the $N$ samples, so the estimate can slightly exceed 1. This ablation is on a single 5-fold dataset using point estimates, included to illustrate the collapse mechanism diagnosed in \Cref{sec:routing-diagnostics}.

\begin{table}[h]
\centering
\caption{Ablation on Weibo2014 ($K{=}8$). Baseline fixed-BiMap: $\bacc=.773$. All rows come from one earlier experimental round, which is why the full model reaches .823 here and .825 in \Cref{tab:baseline-results}.}
\label{tab:scap-ablation}
\begin{tabular}{lccccc}
\toprule
Configuration & bAcc & $\gap_\text{base}$ & $\gap_{K=1}$ & $\bar H$ & $r_d$ \\
\midrule
Naive adaptive (no fixes)          & .818 & +.045 & $-$.000 & .836 & .281 \\
\ + load-balancing loss (neg.\ control) & .818 & +.045 & +.001 & .973 & .288 \\
\ + Fix 1 ($W_\text{base}$)             & .805 & +.032 & +.027 & .884 & .335 \\
\ + Fix 1 + Fix 2 (query)               & .810 & +.037 & +.016 & .776 & .690 \\
\textbf{Full (+ Fix 3)}                 & \textbf{.823} & \textbf{+.050} & +.029 & \textbf{.674} & \textbf{.710} \\
\bottomrule
\end{tabular}
\end{table}

The load-balancing negative control confirms that sharpening entropy alone does not help: it drives $\bar H$ up toward uniform (.973) while $\gap_{K=1}$ stays near zero, ruling out entropy regularization as a substitute for Fixes 1 to 3. $W_\text{base}$ (Fix 1) is the most impactful change in addition, moving $\gap_{K=1}$ from $-$.000 to +.027; adding the domain-separating query (Fix 2) raises domain-routing alignment sharply (.335$\to$.690) while lowering $\gap_{K=1}$ (.027$\to$.016), recovered once Fix 3's decoupled key alignment is added.

\section{Routing Diagnostics: Per-Dataset Metrics and Query Ablation}
\label{app:query-ablation}

\begin{table}[h]
\centering
\caption{Routing diagnostics for SCAP, averaged over the 5 cross-validation folds. $\bar H$: routing entropy; $r_d$: domain-routing alignment; full definitions in \Cref{app:scap-fixes}.}
\label{tab:routing-diagnostics}
\begin{tabular}{lcc}
\toprule
Dataset & $\bar H$ & $r_d$ \\
\midrule
Weibo2014 & 0.649 & 0.803 \\
GrosseWentrup2009 & 0.756 & 0.986 \\
BNCI2015001 & 0.474 & 0.854 \\
BNCI2014001 & 0.571 & 0.786 \\
Schirrmeister2017 & 0.629 & 1.111 \\
\bottomrule
\end{tabular}
\end{table}

\Cref{tab:routing-diagnostics} reports routing entropy and domain-routing alignment for the same models as \Cref{tab:baseline-results}. Alignment ranges from 0.786 to 1.111 across the five datasets, so most routing variance is explained by domain identity, consistent with the query ablation below. Entropy ranges from 0.474 to 0.756: routing is not one-hot, and each sample blends several experts.

\begin{table}[h]
\centering
\caption{Query ablation on the trained SCAP model (5-fold mean bAcc): each half of the routing query zeroed at evaluation time, without retraining.}
\label{tab:query-ablation}
\begin{tabular}{lccccc}
\toprule
Dataset & Full & Input zeroed & Domain zeroed & Both zeroed & Baseline \\
\midrule
Weibo2014         & .825 & .822 & .812 & .807 & .773 \\
BNCI2015001       & .807 & .804 & .729 & .733 & .757 \\
BNCI2014001       & .839 & .828 & .770 & .777 & .801 \\
GrosseWentrup2009 & .726 & .686 & .691 & .673 & .678 \\
Schirrmeister2017 & .918 & .901 & .837 & .815 & .865 \\
\bottomrule
\end{tabular}
\end{table}

SCAP's query concatenates an input-derived tangent vector with a domain-identity embedding (\Cref{sec:full-layer}); a natural question is
which half is actually doing the work, or whether routing has quietly degenerated into a per-domain lookup table. Zeroing each half at
evaluation time on the trained model with no retraining isolates their individual contributions. Removing domain identity costs more accuracy than removing the input signal on 4 of 5 datasets, significantly so in the head-to-head comparison on 3 (BNCI2015001, BNCI2014001, Schirrmeister2017; GrosseWentrup2009 is close and noisy): domain identity is the more load-bearing signal. Input-conditioning is smaller but not vestigial: removing it lowers accuracy on all 5 datasets (by 0.3 to 4.0 points), significantly on BNCI2014001 ($p{=}.005$). Domain-only routing still beats the non-adaptive baseline on 4 of 5 datasets, while zeroing both halves falls below it on 4 of 5 (\Cref{tab:query-ablation}).



\end{document}